\pdfoutput=1
\documentclass[sigconf,authoryear]{acmart}

\copyrightyear{2026}
\acmYear{2026}
\setcopyright{cc}
\setcctype{by}
\acmConference[SA Technical Communications '26]{SIGGRAPH Asia 2026 Technical Communications}{December 01--04, 2026}{Kuala Lumpur, Malaysia}
\acmBooktitle{SIGGRAPH Asia 2026 Technical Communications (SA Technical Communications '26), December 01--04, 2026, Kuala Lumpur, Malaysia}
\acmDOI{10.1145/3829339.3847818}
\acmISBN{979-8-4007-2841-9/2026/12}

\begin{document}


\author{Karthik {Mohan Kumar}}
\orcid{0009-0001-0362-819X}
\affiliation{%
  \institution{Advanced Micro Devices, Inc.}
  \city{San Jose}
  \country{USA}}
\email{Karthik.MohanKumar@amd.com}

\author{Damian Andrysiak}
\orcid{0009-0008-1098-2980}
\affiliation{%
  \institution{Advanced Micro Devices, Inc.}
  \city{{\L}\'od\'z}
  \country{Poland}}
\email{Damian.Andrysiak@amd.com}

\author{Pedro Antonio Pena}
\orcid{0000-0001-9257-5562}
\affiliation{%
  \institution{Advanced Micro Devices, Inc.}
  \city{Orlando}
  \country{USA}}
\email{Pedro.Pena@amd.com}

\author{Kunal Tyagi}
\orcid{0000-0002-4242-1378}
\affiliation{%
  \institution{Advanced Micro Devices, Inc.}
  \city{Munich}
  \country{Germany}}
\email{Kunal.Tyagi@amd.com}

\author{Rama Harihara}
\orcid{0009-0009-5351-8931}
\affiliation{%
  \institution{Advanced Micro Devices, Inc.}
  \city{San Jose}
  \country{USA}}
\email{Rama.Harihara@amd.com}

\renewcommand{\shortauthors}{Mohan Kumar et al.}

\title[DAGS: Disentangled Steering of a Frozen Image DiT]{DAGS: Disentangled Appearance-and-Geometry Steering of a Frozen Image DiT for Temporally Stabilized Generative Rendering}

\begin{abstract}
Diffusion transformers (DiTs) generate high-fidelity images from text and image conditions, but their outputs carry large variance and their faithfulness to a desired target depends heavily on how the condition is supplied. We present \textbf{DAGS}, a lightweight, attention-free, disentangled \emph{appearance} and \emph{geometry} conditioning scheme that steers a \textbf{frozen image DiT} to produce high-fidelity, highly faithful, and \emph{independently controllable} renders. Two small convolutional encoders compute conditioning features \textbf{once per frame} and inject them as a learned, per-layer, element-wise residual into the image tokens, avoiding the quadratic cost of stacking conditions through attention. Because control and temporal handling live \emph{outside} the frozen backbone, we retain its vast pretrained prior and eliminate backbone-overfitting risk. We further add a small recurrent lighting stabilizer and a training-free temporal guidance term that, coupled with our conditioning, elevate a per-frame image model into a streaming renderer. DAGS produces controllable, high-quality renders at a fraction of the compute of path tracing; it is not real-time, trading compute for controllability and quality. On a matched 1-spp\,+\,G-buffer input, \textbf{per-frame} DAGS reconstructs \textbf{+8.6\,dB\,/\,+10.1\,dB PSNR} over the real-time denoiser Intel OIDN and the diffusion renderer RGB$\leftrightarrow$X while being \textbf{2.5--8$\times$ more temporally stable} perceptually (temporal-LPIPS flicker).
\end{abstract}



\begin{teaserfigure}
  \centering
  \includegraphics[width=\textwidth]{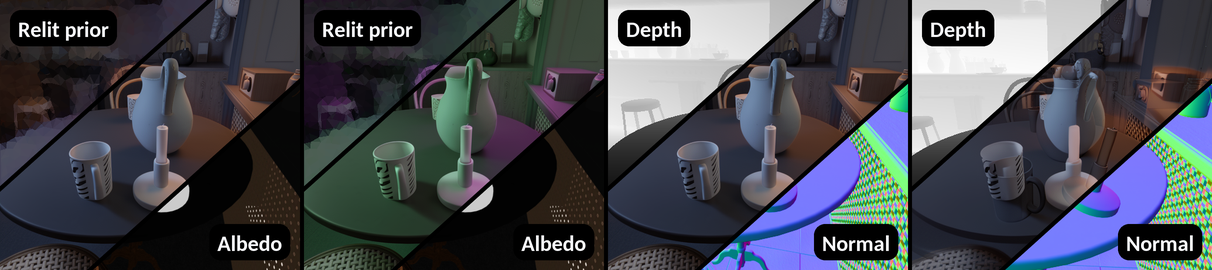}
  \caption{\textbf{Disentangled steering.} Corner triangles show conditioning buffers; the diagonal band shows the DAGS output. \emph{Panels 1--2 (appearance):} inference-time hue rotation $\theta{=}0^\circ$ and $270^\circ$, albedo and geometry fixed; relit prior (top-left) and albedo (bottom-right) bracket the relit output, whose color tracks $\theta$ while structure is unchanged. \emph{Panels 3--4 (geometry):} geometry offsets $+0$ and $+16$, appearance pinned; depth (top-left) and normals (bottom-right) bracket the rendered output, whose structure follows the donor geometry while color stays consistent.}
  \label{fig:teaser}
\end{teaserfigure}

\maketitle

\section{Introduction}
Diffusion transformers~\cite{peebles2023dit} generate high-fidelity images, but \emph{controlling} them precisely remains difficult. The dominant control paradigms---reference-image conditioning, semantic conditioning, and input channel stacking---either operate through the attention stack (whose cost scales quadratically with token count) or require coarse-correcting the signal at many layers to stay faithful. ControlNet~\cite{zhang2023controlnet} achieves strong spatial control but duplicates the encoder ($\approx$\,50\% of the backbone parameters) and re-injects per-layer features, whereas a \emph{small} external encoder can already steer a frozen backbone. Semantic features are not the \emph{metric} geometry a renderer needs, and clip-based video DiTs~\cite{pena2025gfed} add latency without such handles.

We argue that \textbf{entangled control hurts faithfulness}: when a single conditioning stream carries both geometry and appearance, the model is free to be selective about which cue it honors, so structure and color trade off against each other. We instead split control into two disentangled streams and show this improves fidelity and yields independent, predictable control.

\textbf{Concurrent work.} GateControl~\cite{liu2026gatecontrol}, developed independently, reaches a convergent conclusion for \emph{linear-attention} backbones (SANA): a token-wise \emph{gated additive} injection---without multimodal attention and leaving text tokens untouched---is a robust, minimal conditioning mechanism. DAGS differs in three respects: (i) we steer a \emph{full-attention} MMDiT (Flux.2), not a linear-attention model; (ii) we use \emph{two disentangled} encoders for \textbf{independent} geometry-versus-appearance control; and (iii) we target \emph{temporally stabilized generative rendering} from 1-spp\,+\,G-buffers. That the additive, no-cross-attention insight was reached independently on a different backbone corroborates our design choice.

\textbf{Our approach.} DAGS steers a frozen Flux.2-Klein 4\,B DiT with \textbf{$<$\,10\,M trainable parameters} via three choices: \emph{(i)} attention-free, once-per-frame CNN residuals (no token concatenation); \emph{(ii)} independent geometry/appearance control (double dissociation, \S\ref{sec:disent}); \emph{(iii)} temporal stabilization outside the frozen backbone, preserving the full generative prior.

\section{Method}
\begin{figure*}[t]
  \centering
  \includegraphics[width=\textwidth]{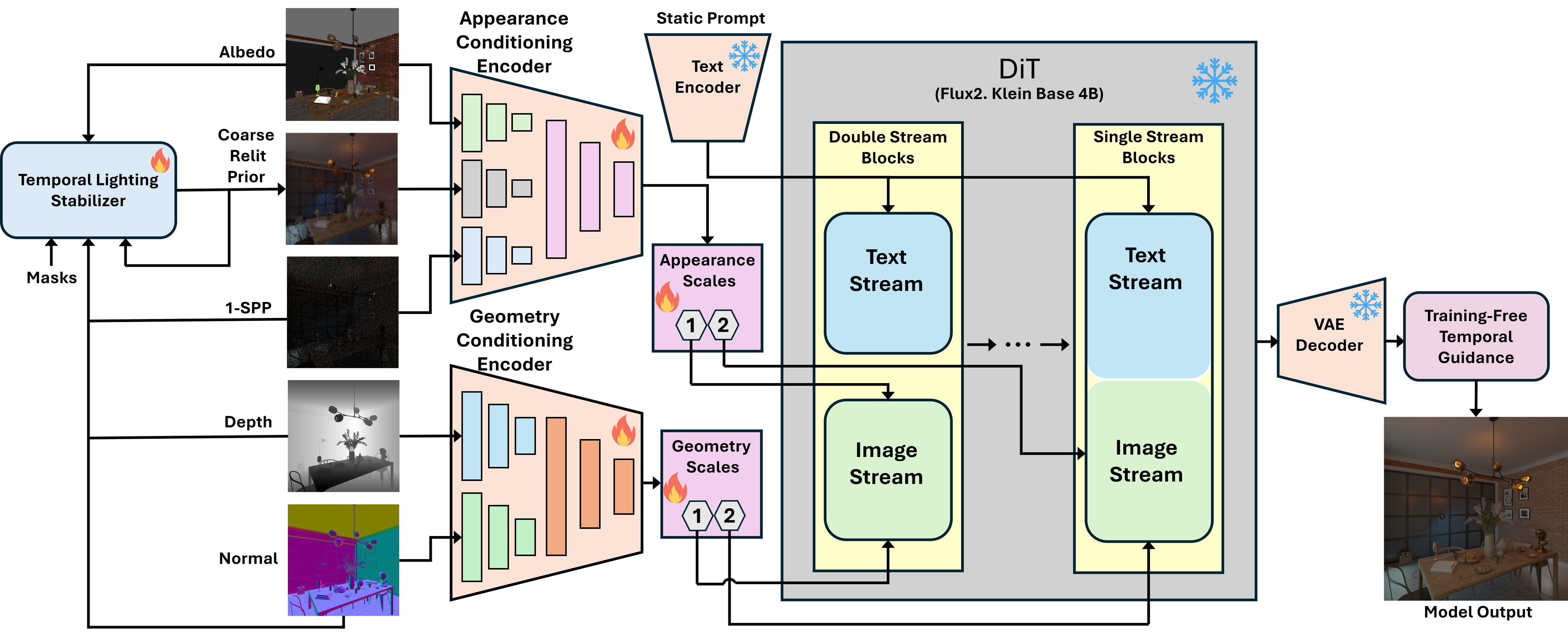}
  \caption{\textbf{DAGS architecture.} DAGS adds two small encoders---one for geometry, one for appearance---that steer a frozen Flux.2-Klein DiT each frame, with a pretrained lighting stabilizer for the relit prior and an optional inference-time step that smooths flicker across frames.}
  \label{fig:arch}
\end{figure*}

\textbf{Overview (Fig.~\ref{fig:arch}).} Given a 1-spp path-traced frame plus per-pixel G-buffers (albedo, depth, normal) and a coarse \emph{relit prior} $R$ (\S\ref{sec:stab}), DAGS predicts the final rendered frame. Two independent encoders---geometry and appearance---encode the buffers into token-aligned features that are scaled per DiT layer and added to the image stream. The DiT, its text encoder (fed a fixed static prompt), and the VAE are all frozen. At test time we use \textbf{ground-truth} G-buffers from the same path tracer as the 1-spp frame (no estimated buffers).

\subsection{Geometry encoder}\label{sec:geom}
Two stems process \textbf{depth} and \textbf{normals} independently before their features are fused and jointly refined into geometry conditioning tokens ($\approx$\,4.80\,M parameters). Channel/feature dropout during training makes the encoder compact and robust to missing or noisy buffers. The geometry stream carries sharp edges, occlusion boundaries, and surface-orientation cues that pin the \emph{structure} of the output while leaving \emph{appearance} free to be inferred by the prior. Because depth and normals are deterministic per frame (no Monte-Carlo noise), the geometry stream is comparatively stable and contributes only a small share of temporal inconsistency.

\subsection{Appearance encoder}
The appearance encoder ($\approx$\,4.82\,M parameters) consumes \textbf{albedo}, the \textbf{raw 1-spp} frame, and the \textbf{coarse relit prior} $R$---a facetized RGB lighting proxy produced upstream by the recurrent stabilizer (\S\ref{sec:stab}). \emph{Raw 1-spp} provides high-frequency lighting---shadows, contact darkening, highlights---that $R$ smooths away. Temporal flicker is dominated by noisy appearance inputs; we stabilize only $R$ (\S\ref{sec:stab}) plus facetization, leaving raw 1-spp unconstrained, so residual specular flicker remains (\S\ref{sec:guidance}, \S\ref{sec:limitations}). Together, raw 1-spp and $R$ carry lighting; albedo adds stable surface-boundary cues (\S\ref{sec:ablation}).

\subsection{Recurrent lighting stabilizer}\label{sec:stab}
A small recurrent U-Net ($\approx$\,0.28\,M parameters) produces the \textbf{relit prior} $R$---a facetized coarse RGB proxy for the fully lit image, built \emph{upstream} from the same 1-spp frame and G-buffers (not a second path-traced pass). We demodulate
\begin{equation*}
L=\log(\mathrm{clip}(I\oslash a)),
\end{equation*}
apply a fixed blur to form a noisy illumination key, and pass it through a U-Net that learns per-frame spatial denoising in log space. A recurrent gate then fuses the backward-warped previous stabilized estimate with the current one---touching \emph{illumination}, not material color:
\begin{equation*}
\hat{L}_t \;=\; L^{\mathrm{s}}_t + v\, g \odot \big(\mathcal{W}(\hat{L}_{t-1}) - L^{\mathrm{s}}_t\big) + \delta,
\end{equation*}
where $\mathcal{W}(\cdot)$ backward-warps by the G-buffer motion field, $L^{\mathrm{s}}_t$ is the spatial (per-frame) estimate, $v$ the validity mask, $g$ the learned gate, and $\delta$ a small residual. Re-modulating gives $a\odot\exp(\hat{L}_t)$, and facetizing yields the relit prior $R$. \textbf{Facetization} Delaunay-triangulates $a\odot\exp(\hat{L}_t)$ into flat color facets, suppressing MC speckle while preserving coarse layout. A validity mask routes disoccluded pixels to the spatial estimate; a specular-motion score (from roughness and motion magnitude) modulates warped history on glossy regions. The stabilizer is trained separately on demodulated lighting sequences and then frozen, while the geometry/appearance encoders are trained later for DiT conditioning. Stabilizing $R$ reduces downstream flicker (Tab.~\ref{tab:temporal}).

\subsection{Conditioning injection}
Each of the \textbf{25 DiT layers} gates once-per-frame token-aligned geometry and appearance encodings
\begin{equation*}
c_g=\mathcal{E}_g(D,N),\qquad c_a=\mathcal{E}_a(A,I_1,R),
\end{equation*}
where $D$, $N$, and $A$ are the depth, normal, and albedo G-buffers and $I_1$ the raw 1-spp RGB frame, with learned per-stream scalars
\begin{equation*}
h^{(\ell)} \;\leftarrow\; h^{(\ell)} + s^{(\ell)}_g\,c_g + s^{(\ell)}_a\,c_a,
\end{equation*}
where $h^{(\ell)}$ is the image-token hidden state at layer $\ell$ and $s^{(\ell)}_g$, $s^{(\ell)}_a$ are the only layer-specific trainable parameters beyond the encoders (25 $\times$ 2 $=$ \textbf{50 scalars}). Injection targets \textbf{image tokens only}, in \textbf{both} double- and single-stream blocks, under a \textbf{static prompt}; the text encoder stays frozen.

\subsection{Flow source}\label{sec:flow}
Because albedo can cover geometry-conditioning failure modes, we replace the flow-matching source~\cite{lipman2023flow} from Gaussian noise with a \textbf{content-carrying} source, learning a shorter transport chord from an approximate image to the fully lit target: we set the source endpoint $x_0=R$ (the relit prior) rather than $x_0\sim\mathcal{N}(0,I)$, shortening the path to the target $x_1$ and lowering the conditional variance the few-step ODE must integrate. The same $R$ conditions the appearance encoder and initializes the flow. Initializing the flow from the \textbf{relit prior} (the shortest chord) beats noise on LPIPS in \textbf{all nine scenes} and on PSNR in 7 of 9 (up to $+8.1$\,dB), and is our \textbf{best model} (Tab.~\ref{tab:flow}). Both learnability and few-step ODE integration error grow with path length and conditional variance; a source that already carries geometry and approximate lighting leaves only a small residual to transport (\S\ref{sec:flowres}).

\subsection{Training-free temporal guidance}\label{sec:guidance}
At inference, we optionally nudge $\hat{x}$ toward the motion-warped previous frame in valid regions:
\begin{equation*}
\hat{x} \;\leftarrow\; \hat{x} + \lambda(\sigma)\, M \odot \big(\mathcal{W}(x_{t-1}) - \hat{x}\big),
\end{equation*}
where $\hat{x}$ is the current decoded estimate, $x_{t-1}$ the previous decoded frame, $\mathcal{W}$ the G-buffer warp from \S\ref{sec:stab}, and $M$ a valid-reprojection mask. The schedule
\begin{equation*}
\lambda(\sigma)=\lambda_{\max}(1-\sigma/\sigma_{\max})_+
\end{equation*}
applies guidance only on low-noise steps ($\lambda_{\max}$, $\sigma_{\max}$ are inference hyperparameters, not learned). This term adds \textbf{no trainable weights}. Combined with the stabilizer it reduces flicker on low/moderate motion (Tab.~\ref{tab:temporal}); on fast \emph{camera} motion reprojection fails, so it is weakened or disabled (\S\ref{sec:temporal}).

\section{Experiments}
We report PSNR / MS-SSIM / LPIPS, CIE-Lab $\Delta$E2000, depth/normal error, and temporal $E_{\text{warp}}$~\cite{lai2018blind} / $t\text{LP}$~\cite{chu2020tecogan}. \S\ref{sec:ablation} validates the two-stream design; \S\ref{sec:disent} tests independent appearance versus\ geometry control (Fig.~\ref{fig:teaser}); later subsections cover flow source, temporal stabilization (Tab.~\ref{tab:ablations}), and OIDN / RGB$\leftrightarrow$X baselines.

\subsection{Geometry versus\ appearance}\label{sec:ablation}
Each stream owns a distinct factor and the combination is best. \textbf{Geometry-only} yields decent depth but the \textbf{worst (or near-worst) color} (Lab $\Delta$E $6.9\!\rightarrow\!21.5$ if harder; worst in 8 of 9 scenes); \textbf{appearance-only} yields good color but the \textbf{worst geometry} (depth error up to ${\sim}0.6$); \textbf{geometry+appearance inherits both}, taking best/tied MS-SSIM+LPIPS+geometry in all nine scenes and best PSNR in 8 of 9 scenes. The two factors are thus \textbf{complementary and non-competing}: adding either stream leaves the other's axis intact, so the design captures both without a structure-versus-color compromise. We also ablate albedo \textbf{within the appearance-only model} (geometry encoder disabled): with albedo, inferred depth error falls in \textbf{7 of 9} scenes---often by half---while Lab $\Delta$E stays flat. Albedo therefore supplies \textbf{boundary-aligned structure cues} to the appearance stream rather than acting as a relighting knob.

\subsection{Disentanglement}\label{sec:disent}
\textbf{Inference controls.} We edit the \emph{raw} G-buffer inputs before encoding, so conditioning and paired GT stay consistent. \emph{Lighting:} hue-rotate the lit renders that feed the relit branch (albedo/geometry fixed). \emph{Geometry:} pin appearance to reference frame $t$ and swap in donor depth+normals from frame $t{+}\Delta$ (reference albedo, 1-spp, and $R$ unchanged), sweeping $\Delta$.

\textbf{Appearance.} \textit{Hue-response} is the slope of generated versus\ GT local hue rotation across the sweep ($\approx 1$ = faithful relighting), and \textit{hue-match} is cosine similarity of hue per-pixel ($1$ = perfect local color agreement). Rotating the hue of the \emph{lighting} conditioning (Fig.~\ref{fig:teaser}, panels 1--2) tracks the relit target---hue-response $\approx$\,0.93 (0.89--0.99) and hue-match $\approx$\,0.85 across eight scenes---while monocular geometry error stays flat (AbsRel spread $\le 0.11$, normal spread $\le 2.8^\circ$). A \textbf{negative control} that rotates only conditioning \emph{albedo} is correctly \emph{rejected} (hue-response $\approx 0$), confirming the model honors lighting, not arbitrary material edits.

\textbf{Geometry.} \textit{Monocular geometry error} is depth (AbsRel) and normal (angular) error inferred from the rendered RGB with Depth Anything V2~\cite{yang2024dav2} and DSINE~\cite{bae2024dsine}, compared with the same estimators on GT (flat across the sweep = structure unchanged). Pinning appearance and swapping donor depth+normals from $t{+}\Delta$, inferred depth in the output shifts toward the donor geometry when the reference and donor frames differ enough in layout---strongest on three high-motion scenes ($f_{\text{depth}}=0.84/0.81/0.71$), weaker on low-motion interiors (median $0.57$ over nine scenes); Fig.~\ref{fig:teaser} panels 3--4 show the corresponding depth/normal conditioning at donor offsets $+0$ and $+16$. We score this with
\begin{equation*}
f_{\text{depth}}=v_{\text{o}}/(v_{\text{o}}+v_{\text{d}})
\end{equation*}
(origin, donor) from monocular depth at the largest $\Delta$ ($1$ = follows donor, $0.5$ = inconclusive). \textbf{Color stays fixed:} Lab $\Delta$E spread $\le 0.73$ and hue slope $\le 0.06$ in 8 of 9 scenes (geometry edits do not relight).

\textbf{Mechanistic probe.} Comparing the geometry and appearance residual embeddings with CKA / RBF-CKA / Gaussian-MI (9 scenes), the raw similarity is almost entirely the trivial \textbf{positional} (shared token-grid) subspace. Once position is removed, cross-branch \textbf{content} similarity is small (CKA $\le 0.32$) and far below within-branch structure (0.95--0.99), though above the shuffled null. We thus claim the branches are \textbf{disentangled up to a shared positional subspace, with small residual content overlap}---this supports rather than overclaims the behavioral result.

\subsection{Flow source}\label{sec:flowres}
We compare three \emph{training-time} flow sources: Gaussian noise, albedo, and the relit prior $R$. Each is a separate two-stream checkpoint evaluated with \textbf{non-temporal} inference (Tab.~\ref{tab:flow}). \textbf{Relit-flow} is strongest: best LPIPS in \textbf{all nine} scenes and best/near-best PSNR in 7 of 9 (mean $+2.0$\,dB over noise). Gains are largest where the transport residual is hardest---up to $+8.1$\,dB on the hardest cases---because $R$ already carries approximate geometry and lighting, shortening the flow chord. \textbf{Albedo-flow} is conditional: it can slightly beat noise on easy interiors but \textbf{collapses} on hard cases (up to $-5.0$\,dB) when the shading gap between albedo and shadowed GT is large. We therefore report relit-flow $\gg$ noise as the robust claim.

\subsection{Temporal stabilization}\label{sec:temporal}
On the fixed relit-flow model we factorially ablate two inference-time mechanisms: recurrent lighting stabilizer $P$ (upstream on $R$) and training-free latent guidance $G$ (downstream on decoded frames). Tab.~\ref{tab:temporal} reports GT-referenced temporal metrics averaged over nine scenes: $E_{\text{warp}}$ (flow-warped pixel error) and $t\text{LP}$ (unwarped consecutive-frame LPIPS).

\textbf{Mean over nine scenes ($P$ and $G$ both on).} Compared with per-frame inference without either mechanism, enabling both reduces mean $E_{\text{warp}}$ from $3.42$ to $2.52$ and mean $t\text{LP}$ from $0.0115$ to $0.0053$---a $2.2\times$ drop in perceptual flicker. The effects are additive: $P$ alone cuts mean $t\text{LP}$ by $41\%$ by stabilizing the relit prior; $G$ alone improves decoded temporal coherence; together they compound.

\textbf{Per-scene reductions (versus\ per-frame baseline).} Raw $E_{\text{warp}}$ falls most on static, low-motion content ($\mathbf{-77\%}$), and by $-31\%$ to $-41\%$ on moderate object or interior motion, using each scene's best-performing configuration from the factorial (typically $P{+}G$). On fast camera motion, reprojection fails, gains vanish ($\approx 0\%$), and we disable $P$, using only low-$\lambda$ $G$ or per-frame inference.

\begin{table}[t]
\centering
\small
\setlength{\tabcolsep}{2pt}
\caption{Flow and Temporal ablations (mean over nine scenes). \textbf{Left:} flow $x_0$ (non-temporal). \textbf{Right:} temporal arms on relit-flow; $P$ on $R$, $G$ on latents.}
\label{tab:ablations}
\label{tab:flow}
\label{tab:temporal}
\begin{minipage}[t]{0.58\columnwidth}
\centering
{\footnotesize\textit{Flow source (no exposure gain)}}\\[0.25em]
\begin{tabular*}{\linewidth}{@{\extracolsep{\fill}}lccc@{}}
\toprule
$x_0$ & PSNR\,$\uparrow$ & MS-SSIM\,$\uparrow$ & LPIPS\,$\downarrow$ \\
\midrule
Noise              & 26.2 & .919 & .149 \\
Albedo             & 25.6 & .912 & .151 \\
\textbf{Relit $R$} & \textbf{28.2} & \textbf{.933} & \textbf{.107} \\
\bottomrule
\end{tabular*}
\end{minipage}\hfill
\begin{minipage}[t]{0.38\columnwidth}
\centering
{\footnotesize\textit{Temporal stabilization}}\\[0.25em]
\begin{tabular*}{\linewidth}{@{\extracolsep{\fill}}lcc@{}}
\toprule
Arm & $E_{\text{warp}}$\,$\downarrow$ & $t\text{LP}$\,$\downarrow$ \\
\midrule
none & 3.42 & .0115 \\
$P$  & 2.74 & .0068 \\
$G$  & 2.84 & .0080 \\
\textbf{$G{+}P$} & \textbf{2.52} & \textbf{.0053} \\
\bottomrule
\end{tabular*}
\end{minipage}
\end{table}

\subsection{Comparison}
\begin{figure*}[t]
  \centering
  \includegraphics[width=0.92\textwidth]{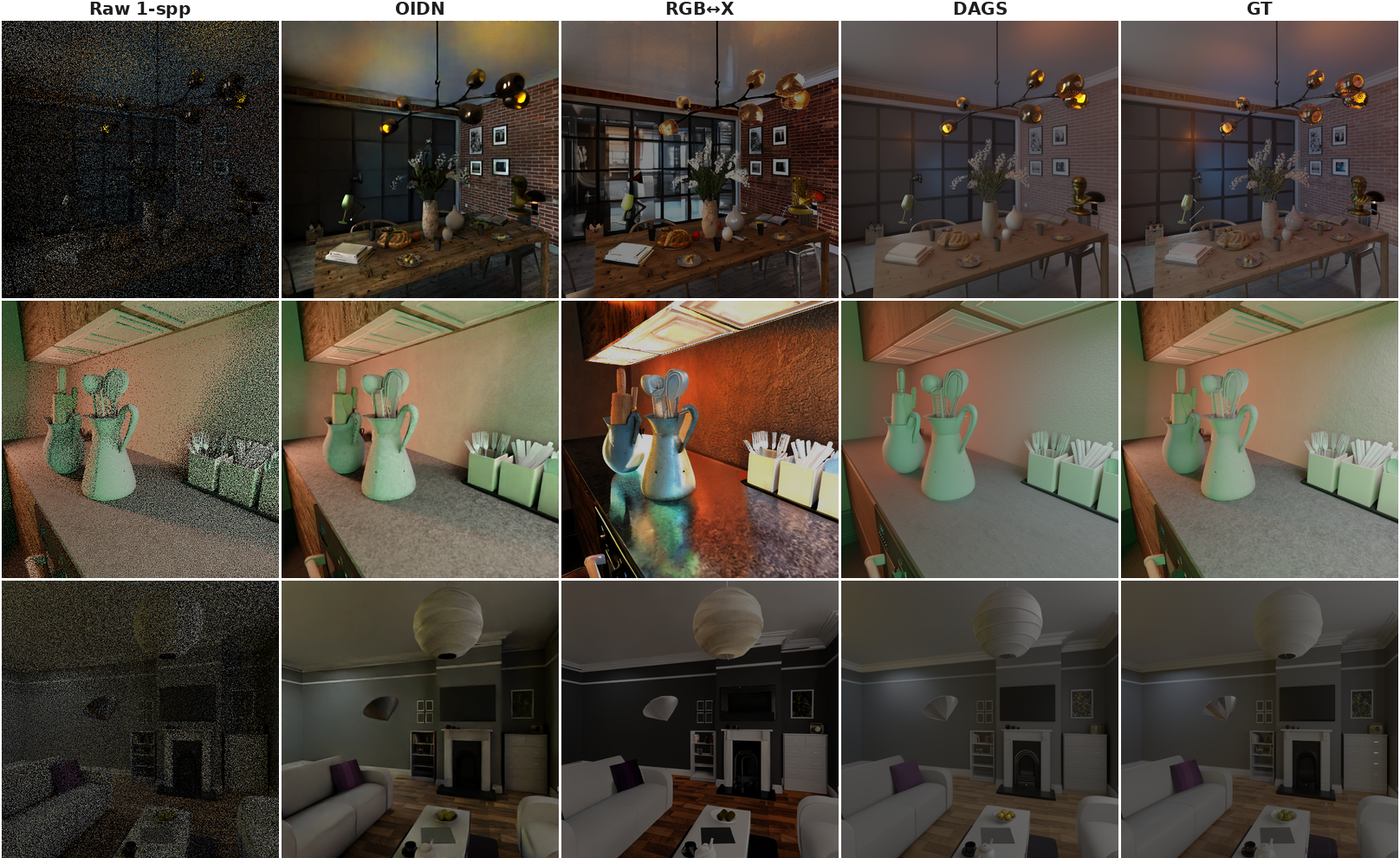}
  \caption{\textbf{Reconstruction vs.\ baselines} (exposure-matched OIDN/RGB$\leftrightarrow$X). Three held-out scenes; DAGS ($G{+}P$) is visually closest to GT on both detail and color, while OIDN blurs structure and RGB$\leftrightarrow$X drifts in color and flickers temporally (Tab.~\ref{tab:baselines}).}
  \label{fig:baselines}
\end{figure*}

On the same 1-spp\,+\,G-buffer budget we compare to \textbf{OIDN}~\cite{afra2019oidn} (training-free denoiser; reconstruction axis) and \textbf{RGB$\leftrightarrow$X} \cite{zeng2024rgbx} (SD1.5 forward renderer; controllability axis). Tab.~\ref{tab:baselines} and Fig.~\ref{fig:baselines} report quantitative and qualitative results; numbers are scale-invariant (a per-frame gain removes an exposure confound on reconstruction; temporal metrics match Tab.~\ref{tab:temporal}, \S\ref{sec:temporal}; raw numbers widen our lead).

\begin{table}[t]
\centering
\small
\setlength{\tabcolsep}{2pt}
\caption{DAGS versus\ baselines, mean over nine scenes, matched inputs and GT. Reconstruction uses a per-frame exposure gain; temporal metrics match Tab.~\ref{tab:temporal} (\S\ref{sec:temporal}).}
\label{tab:baselines}
\begin{tabular*}{\columnwidth}{@{\extracolsep{\fill}}lccccc@{}}
\toprule
Method & PSNR\,$\uparrow$ & MS-SSIM\,$\uparrow$ & LPIPS\,$\downarrow$ & $t\text{LP}$\,$\downarrow$ & $E_{\text{warp}}$\,$\downarrow$ \\
\midrule
raw 1-spp (floor)        & 13.70 & ---    & ---    & ---    & ---   \\
OIDN (denoiser)          & 21.23 & 0.8159 & 0.2219 & 0.0285 & 3.17 \\
RGB$\leftrightarrow$X    & 19.70 & 0.7028 & 0.3200 & 0.0427 & 9.14 \\
\textbf{DAGS (per-frame)} & \textbf{29.79} & \textbf{0.9363} & \textbf{0.1042} & 0.0115 & 3.42 \\
\textbf{DAGS (stab., $G{+}P$)} & 29.39 & 0.9359 & 0.1114 & \textbf{0.0053} & \textbf{2.52} \\
\bottomrule
\end{tabular*}
\end{table}

DAGS wins reconstruction on \textbf{9 of 9 scenes} versus\ both baselines ($+8.6$\,dB over OIDN, $+10.1$\,dB over RGB$\leftrightarrow$X, $\sim$2--3$\times$ lower LPIPS) and perceptual stability ($t\text{LP}$) on \textbf{9 of 9} ($2.5\times$ lower than OIDN per-frame, $5.3\times$ stabilized). Stabilization ($G{+}P$) trades a small amount of per-frame fidelity for markedly lower flicker. OIDN cannot relight/edit; RGB$\leftrightarrow$X adds controllability but, as a per-frame model, flickers badly ($t\text{LP}$ $\sim$8$\times$ worse than DAGS-stab) even when fed the \emph{same} stabilized lighting, and drifts from GT---and its full intrinsic stack (extra roughness and metallic channels that DAGS never sees) changes nothing ($-0.2$\,dB). This is a matched-\emph{input-budget} comparison, not matched compute: OIDN is the training-free quality reference, not a speed competitor.

\section{Limitations and future work}\label{sec:limitations}
DAGS has three main limitations, each pointing to a concrete next step. \textbf{(i) Domain gap:} training uses \textbf{indoor scenes only}; we evaluate on held-out \emph{and} out-of-distribution scenes, but fidelity can drop far from the training distribution---a larger, more diverse corpus is the remedy. \textbf{(ii) Residual flicker:} \textbf{high-frequency detail and specular highlights still flicker} where the appearance inputs are themselves frame-to-frame inconsistent, motivating detail- and specular-aware appearance conditioning. \textbf{(iii) Motion-regime dependence:} under fast \textbf{camera} motion, reprojection validity is low, so guidance cannot beat per-frame inference and the recurrent stabilizer must be disabled to avoid ghosting. Our temporal handling is \textbf{external} to the frozen backbone (input-space stabilizer + output-space guidance), so it relies on reprojection that breaks under heavy disocclusion; folding temporal feedback into diffusion sampling---conditioning each step on the previous decoded frame while keeping the backbone frozen---could raise this ceiling.

\begin{acks}
We thank \citet{anand2026temporally} for sharing the datasets generated for their work, and our colleagues at Advanced Micro Devices, Inc., whose contributions made this research possible.
\end{acks}

\bibliographystyle{ACM-Reference-Format}
\bibliography{references_arxiv}

@inproceedings{anand2026temporally,
  author = {Anand, Harish and Kuznetsov, Alexandr and Kim, Sungye and Uss, Wojciech and Kali\'nski, Wojciech and Harihara, Rama},
  title = {Temporally Stable Generative Illumination with a One-Step Diffusion Model},
  booktitle = {Computer Vision -- {ECCV} 2026},
  pages = {205--224},
  year = {2026},
  publisher = {Springer},
  doi = {10.1007/978-3-032-37026-6_12},
}

@misc{afra2019oidn,
  author = {\'Afra, Attila T.},
  title = {Intel {Open Image Denoise}},
  year = {2019},
  howpublished = {\url{https://openimagedenoise.github.io}},
}

@inproceedings{bae2024dsine,
  author = {Bae, Gwangbin and Davison, Andrew J.},
  title = {Rethinking Inductive Biases for Surface Normal Estimation},
  booktitle = {{IEEE}/{CVF} Conference on Computer Vision and Pattern Recognition ({CVPR})},
  pages = {9535--9545},
  year = {2024},
  doi = {10.1109/CVPR52733.2024.00911},
}

@article{chu2020tecogan,
  author = {Chu, Mengyu and Xie, You and Mayer, Jonas and Leal-Taix\'e, Laura and Thuerey, Nils},
  title = {Learning Temporal Coherence via Self-Supervision for {GAN}-based Video Generation},
  journal = {{ACM} Trans. Graph.},
  volume = {39},
  number = {4},
  articleno = {75},
  numpages = {19},
  year = {2020},
  doi = {10.1145/3386569.3392457},
}

@inproceedings{lai2018blind,
  author = {Lai, Wei-Sheng and Huang, Jia-Bin and Wang, Oliver and Shechtman, Eli and Yumer, Ersin and Yang, Ming-Hsuan},
  title = {Learning Blind Video Temporal Consistency},
  booktitle = {Computer Vision -- {ECCV} 2018},
  pages = {170--185},
  year = {2018},
  publisher = {Springer},
  doi = {10.1007/978-3-030-01219-9_11},
}

@inproceedings{lipman2023flow,
  author = {Lipman, Yaron and Chen, Ricky T. Q. and Ben-Hamu, Heli and Nickel, Maximilian and Le, Matt},
  title = {Flow Matching for Generative Modeling},
  booktitle = {International Conference on Learning Representations ({ICLR})},
  year = {2023},
}

@inproceedings{liu2026gatecontrol,
  author = {Liu, Yuhe and Tan, Zhenxiong and Hu, Yujia and Liu, Songhua and Wang, Xinchao},
  title = {Gated Condition Injection without Multimodal Attention: Towards Controllable Linear-Attention Transformers},
  booktitle = {{IEEE}/{CVF} Conference on Computer Vision and Pattern Recognition ({CVPR})},
  pages = {23388--23397},
  year = {2026},
}

@inproceedings{peebles2023dit,
  author = {Peebles, William and Xie, Saining},
  title = {Scalable Diffusion Models with Transformers},
  booktitle = {{IEEE}/{CVF} International Conference on Computer Vision ({ICCV})},
  pages = {4172--4182},
  year = {2023},
  doi = {10.1109/ICCV51070.2023.00389},
}

@inproceedings{pena2025gfed,
  author = {Pena, Pedro Antonio and {Mohan Kumar}, Karthik and Andrysiak, Damian and Tyagi, Kunal and Harihara, Rama},
  title = {{G-FED}: {G}-Buffer Guided Frame Extrapolation in Video Diffusion Models},
  booktitle = {{ACM SIGGRAPH} 2025 Posters},
  year = {2025},
  doi = {10.1145/3721250.3742972},
}

@inproceedings{yang2024dav2,
  author = {Yang, Lihe and Kang, Bingyi and Huang, Zilong and Zhao, Zhen and Xu, Xiaogang and Feng, Jiashi and Zhao, Hengshuang},
  title = {Depth Anything {V2}},
  booktitle = {Advances in Neural Information Processing Systems ({NeurIPS})},
  year = {2024},
}

@inproceedings{zeng2024rgbx,
  author = {Zeng, Zheng and Deschaintre, Valentin and Georgiev, Iliyan and Hold-Geoffroy, Yannick and Hu, Yiwei and Luan, Fujun and Yan, Ling-Qi and Ha\v{s}an, Milo\v{s}},
  title = {{RGB}$\leftrightarrow${X}: Image Decomposition and Synthesis Using Material- and Lighting-aware Diffusion Models},
  booktitle = {{ACM SIGGRAPH} 2024 Conference Papers},
  year = {2024},
  doi = {10.1145/3641519.3657445},
}

@inproceedings{zhang2023controlnet,
  author = {Zhang, Lvmin and Rao, Anyi and Agrawala, Maneesh},
  title = {Adding Conditional Control to Text-to-Image Diffusion Models},
  booktitle = {{IEEE}/{CVF} International Conference on Computer Vision ({ICCV})},
  pages = {3813--3824},
  year = {2023},
  doi = {10.1109/ICCV51070.2023.00367},
}

\end{document}